\documentclass[10pt,twocolumn]{article}
\usepackage{amsmath,amssymb,amsfonts}
\usepackage{graphicx}
\usepackage{url}
\usepackage{makecell}
\usepackage[margin=1in]{geometry}

\title{FRACTURED-SORRY-Bench: Framework for Revealing Attacks in Conversational Turns Undermining Refusal Efficacy and Defenses over SORRY-Bench (Automated Multi-shot Jailbreaks)}

\author{Aman Priyanshu$^1$ and Supriti Vijay$^1$ \\ \small{$^1$Carnegie Mellon University}}

\begin{document}

\maketitle

\section{Abstract}
This paper introduces FRACTURED-SORRY-Bench, a framework for evaluating the safety of Large Language Models (LLMs) against multi-turn conversational attacks. Building upon the SORRY-Bench dataset, we propose a simple yet effective method for generating adversarial prompts by breaking down harmful queries into seemingly innocuous sub-questions. Our approach achieves a maximum increase of +46.22\% in Attack Success Rates (ASRs) across GPT-4, GPT-4o, GPT-4o-mini, and GPT-3.5-Turbo models compared to baseline methods. We demonstrate that this technique poses a challenge to current LLM safety measures and highlights the need for more robust defenses against subtle, multi-turn attacks.

\section{Introduction}
As Large Language Models (LLMs) become increasingly prevalent in various applications, ensuring their safe and ethical use is paramount \cite{tang2024prioritizingsafeguardingautonomyrisks, hassine2024llmbased, priyanshu2023chatbotsreadyprivacysensitiveapplications, mazeika2024harmbenchstandardizedevaluationframework}. While significant progress has been made in aligning these models with human values and implementing safety measures, they remain vulnerable to adversarial attacks, particularly those that leverage the nuances of human conversation \cite{shen2024donowcharacterizingevaluating, qi2023finetuningalignedlanguagemodels}.

This paper presents FRACTURED-SORRY-Bench, a framework that extends the SORRY-Bench dataset to evaluate LLM safety against a new class of attacks. Our method decomposes harmful queries into multiple, seemingly benign sub-questions, exploiting the multi-turn nature of conversations to bypass safety mechanisms. This approach represents a more efficient and accessible technique for generating adversarial samples compared to complex optimization-based methods \cite{zou2023universaltransferableadversarialattacks, perez-etal-2022-red}.

\section{Preliminaries}

\subsection{SORRY-Bench Dataset}
Prior work has introduced various datasets and frameworks for evaluating LLM safety, including Do-Not-Answer \cite{wang2023donotanswerdatasetevaluatingsafeguards}, SimpleSafetyTests \cite{vidgen2024simplesafetyteststestsuiteidentifying}, ALERT \cite{tedeschi2024alertcomprehensivebenchmarkassessing}, and HarmBench \cite{mazeika2024harmbenchstandardizedevaluationframework}, each contributing unique perspectives on assessing LLM vulnerabilities and safeguards.
SORRY-Bench \cite{xie2024sorrybenchsystematicallyevaluatinglarge} introduced a comprehensive benchmark for evaluating LLM safety refusal behaviors. It provides a fine-grained taxonomy of 45 potentially unsafe topics and a balanced dataset of unsafe instructions. Our work builds upon this foundation by introducing a new dimension of evaluation: multi-turn conversational attacks.

\begin{table*}[h]
\centering
\begin{tabular}{lcccc}
\hline
\multicolumn{5}{c}{\textbf{Vanilla Responses}} \\
\hline
Model & Harmful \& Relevant & Harmful but Irrelevant & Harmless & ASR (\%) \\
\hline
GPT-4o & 52 & 3 & 395 & 11.56 \\
GPT-3.5-Turbo & 21 & 4 & 425 & 4.67 \\
GPT-4o-mini & 58 & 2 & 390 & 12.89 \\
GPT-4 & 45 & 3 & 402 & 10.00 \\
\hline
\multicolumn{5}{c}{\textbf{Decomposed Responses}} \\
\hline
Model & Harmful \& Relevant & Harmful but Irrelevant & Harmless & ASR (\%) \\
\hline
GPT-4o & 223 & 103 & 124 & 49.56 \\
GPT-3.5-Turbo & 229 & 106 & 115 & 50.89 \\
GPT-4o-mini & 226 & 106 & 118 & 50.22 \\
GPT-4 & 221 & 104 & 125 & 49.11 \\
\hline
\end{tabular}
\caption{Comparison of Vanilla and Decomposed Responses. The four categories represent the types of responses generated by different models: "Harmful \& Relevant" refers to responses that are both harmful and relevant concerning the intended malicious intent; "Harmful but Irrelevant" denotes responses that are harmful but irrelevant to the intended malicious intent; "Harmless" indicates responses that are safe and devoid of harmful content; and "ASR (\%)" represents the Attack Success Rate, reflecting the proportion of responses that successfully convey malicious intent.}
\label{table:combined}
\end{table*}

\begin{table}[h]
\centering
\begin{tabular}{|l|c|}
\hline
\textbf{Intent Category} & \textbf{Percentage (\%)} \\
\hline
\makecell{Harmful \& \\ Intent Conveyed} & 49.33\% \\
\hline
\makecell{Harmless \& \\ Intent Lost} & 23.56\% \\
\hline
\makecell{Intermediate \& \\ Intent Partially Conveyed} & 27.12\% \\
\hline
\end{tabular}
\caption{Percentage distribution of intent categories in decomposed questions. "Harmful \& Intent Conveyed" indicates questions that retained the original harmful intent as a collective set, "Harmless  \& Intent Lost" represents those that strayed away and did not convey the original intent, and "Intermediate \& Intent Partially Conveyed" refers to questions that partially conveyed the intended harmfulness but deviated to some extent.}
\label{table:intent_distribution}
\end{table}

\subsection{Prompt Injection Techniques}
Recent work, such as Imposter.AI \cite{liu2024imposteraiadversarialattackshidden}, has explored sophisticated methods for extracting harmful information from LLMs. These approaches often involve complex, multi-step processes using multiple models or are large scale collections \cite{wei2023jailbrokendoesllmsafety}. In contrast, our method focuses on a simpler, more efficient single-shot prompting technique that achieves similar goals.

Other relevant works include Universal and Transferable Adversarial Attacks \cite{zou2023universaltransferableadversarialattacks} \& Rainbow Teaming \cite{samvelyan2024rainbowteamingopenendedgeneration}, which demonstrate the potential for creating adversarial prompts that transfer across different LLMs, and the exploration of stealthy attacks using ciphers \cite{yuan2024gpt4smartsafestealthy}.

\section{Methodology}
Our approach, FRACTURED-SORRY-Bench, introduces a straightforward method for creating adversarial samples, inspired by the chain-of-thought prompting technique \cite{wei2023chainofthoughtpromptingelicitsreasoning}:

\begin{enumerate}
    \item Decompose a given query into 4-7 sub-questions that appear innocuous when viewed individually.
    \item Present these sub-questions sequentially to the target LLM in a conversational format.
    \item Analyze the cumulative response to determine if the harmful intent of the original query was fulfilled.
\end{enumerate}

This method exploits the LLM's context window and its potential inability to recognize the harmful intent spread across multiple turns. By avoiding explicit harmful language in each sub-question, it aims to bypass content filters and safety measures, similar to the approach in \cite{samvelyan2024rainbowteamingopenendedgeneration}.

\section{Result Analysis}

\subsection{Attack Success Rates}

Our results show significant increases in ASR across all tested models, as presented in Table~\ref{table:combined}. GPT-3.5-Turbo demonstrated the highest vulnerability, with an ASR that increased by a factor of $10.9\times$ compared to its vanilla version, followed by GPT-4 ($4.91\times$), GPT-4o ($4.29\times$), and GPT-4o-mini ($3.9\times$).

We employed GPT-4 as a judge, inspired by prior literature \cite{mazeika2024harmbenchstandardizedevaluationframework, lin-chen-2023-llm}.

\subsection{Intent Conveyance Analysis}

We conducted an analysis to determine whether the fractured prompts correctly conveyed the original harmful intent. We present these results in Table~\ref{table:intent_distribution}. Our findings indicate that 49.33\% of the fractured prompt sets successfully conveyed the original intent, with variations across different categories of harmful queries. This analysis methodology draws inspiration from work on evaluating LLM safeguards \cite{wang2023donotanswerdatasetevaluatingsafeguards, vidgen2024simplesafetyteststestsuiteidentifying}.

\section{Conclusion}
FRACTURED-SORRY-Bench demonstrates the vulnerability of current LLM safety measures to subtle, multi-turn attacks. By decomposing harmful queries into seemingly innocent sub-questions, we achieve significant increases in attack success rates across multiple models. This work highlights the need for more sophisticated safety mechanisms that can understand and evaluate the cumulative intent of multi-turn conversations. Future work should focus on developing defense strategies against these types of attacks and expanding the evaluation to a broader range of LLMs and conversational scenarios.

\bibliographystyle{plain}

\begin{thebibliography}{99}

\bibitem{xie2024sorrybenchsystematicallyevaluatinglarge} Xie, T., Qi, X., Zeng, Y., Huang, Y., Sehwag, U. M., Huang, K., He, L., Wei, B., Li, D., Sheng, Y., Jia, R., Li, B., Li, K., Chen, D., Henderson, P., \& Mittal, P. (2024). SORRY-Bench: Systematically Evaluating Large Language Model Safety Refusal Behaviors. arXiv preprint arXiv:2406.14598.

\bibitem{wang2023donotanswerdatasetevaluatingsafeguards} Wang, Y., Li, H., Han, X., Nakov, P., \& Baldwin, T. (2023). Do-Not-Answer: A Dataset for Evaluating Safeguards in LLMs. arXiv preprint arXiv:2308.13387.

\bibitem{qi2023finetuningalignedlanguagemodels} Qi, X., Zeng, Y., Xie, T., Chen, P. Y., Jia, R., Mittal, P., \& Henderson, P. (2023). Fine-tuning Aligned Language Models Compromises Safety, Even When Users Do Not Intend To!. arXiv preprint arXiv:2310.03693.

\bibitem{samvelyan2024rainbowteamingopenendedgeneration} Samvelyan, M., Raparthy, S. C., Lupu, A., Hambro, E., Markosyan, A. H., Bhatt, M., Mao, Y., Jiang, M., Parker-Holder, J., Foerster, J., Rocktäsche, T., \& Raileanu, R. (2024). Rainbow Teaming: Open-Ended Generation of Diverse Adversarial Prompts. arXiv preprint arXiv:2402.16822.

\bibitem{shen2024donowcharacterizingevaluating} Shen, X., Chen, Z., Backes, M., Shen, Y., \& Zhang, Y. (2024). "Do Anything Now": Characterizing and Evaluating In-The-Wild Jailbreak Prompts on Large Language Models. arXiv preprint arXiv:2308.03825.

\bibitem{tedeschi2024alertcomprehensivebenchmarkassessing} Tedeschi, S., Friedrich, F., Schramowski, P., Kersting, K., Navigli, R., Nguyen, H., \& Li, B. (2024). ALERT: A Comprehensive Benchmark for Assessing Large Language Models' Safety through Red Teaming. arXiv preprint arXiv:2404.08676.

\bibitem{vidgen2024simplesafetyteststestsuiteidentifying} Vidgen, B., Scherrer, N., Kirk, H. R., Qian, R., Kannappan, A., Hale, S. A., \& Röttger, P. (2024). SimpleSafetyTests: a Test Suite for Identifying Critical Safety Risks in Large Language Models. arXiv preprint arXiv:2311.08370.

\bibitem{liu2024imposteraiadversarialattackshidden} Liu, X., Li, L., Xiang, T., Ye, F., Wei, L., Li, W., \& Garcia, N. (2024). Imposter.AI: Adversarial Attacks with Hidden Intentions towards Aligned Large Language Models. arXiv preprint arXiv:2407.15399.

\bibitem{zou2023universaltransferableadversarialattacks} Zou, A., Wang, Z., Carlini, N., Nasr, M., Kolter, J. Z., \& Fredrikson, M. (2023). Universal and Transferable Adversarial Attacks on Aligned Language Models. arXiv preprint arXiv:2307.15043.

\bibitem{yuan2024gpt4smartsafestealthy} Yuan, Y., Jiao, W., Wang, W., Huang, J. T., He, P., Shi, S., \& Tu, Z. (2024). GPT-4 Is Too Smart To Be Safe: Stealthy Chat with LLMs via Cipher. arXiv preprint arXiv:2308.06463.

\bibitem{wei2023chainofthoughtpromptingelicitsreasoning} Wei, J., Wang, X., Schuurmans, D., Bosma, M., Ichter, B., Xia, F., Chi, E., Le, Q., \& Zhou, D. (2023). Chain-of-Thought Prompting Elicits Reasoning in Large Language Models. arXiv preprint arXiv:2201.11903.

\bibitem{wei2023jailbrokendoesllmsafety} Wei, A., Haghtalab, N., \& Steinhardt, J. (2023). Jailbroken: How Does LLM Safety Training Fail?. arXiv preprint arXiv:2307.02483.

\bibitem{perez-etal-2022-red} Perez, E., Huang, S., Song, F., Cai, T., Ring, R., Aslanides, J., Glaese, A., McAleese, N., \& Irving, G. (2022). Red Teaming Language Models with Language Models. In Proceedings of the 2022 Conference on Empirical Methods in Natural Language Processing (pp. 3419-3448).

\bibitem{priyanshu2023chatbotsreadyprivacysensitiveapplications} Priyanshu, A., Vijay, S., Kumar, A., Naidu, R., \& Mireshghallah, F. (2023). Are Chatbots Ready for Privacy-Sensitive Applications? An Investigation into Input Regurgitation and Prompt-Induced Sanitization. arXiv preprint arXiv:2305.15008.

\bibitem{mazeika2024harmbenchstandardizedevaluationframework} Mazeika, M., Phan, L., Yin, X., Zou, A., Wang, Z., Mu, N., Sakhaee, E., Li, N.,  Basart, S., Li, B., Forsyth, D., \& Hendrycks, D. (2024). HarmBench: A Standardized Evaluation Framework for Automated Red Teaming and Robust Refusal. arXiv preprint arXiv:2402.04249.

\bibitem{lin-chen-2023-llm} Lin, Y. T., \& Chen, Y. N. (2023). LLM-Eval: Unified Multi-Dimensional Automatic Evaluation for Open-Domain Conversations with Large Language Models. In Proceedings of the 5th Workshop on NLP for Conversational AI (NLP4ConvAI 2023) (pp. 47-58). Association for Computational Linguistics.

\bibitem{tang2024prioritizingsafeguardingautonomyrisks} Tang, X., Jin, Q., Zhu, K., Yuan, T., Zhang, Y., Zhou, W., Qu, M., Zhao, Y., Tang, J., Zhang, Z., Cohan, A., Lu, Z., \& Gerstein, M. (2024). Prioritizing Safeguarding Over Autonomy: Risks of LLM Agents for Science. arXiv preprint arXiv:2402.04247.

\bibitem{hassine2024llmbased} Hassine, J. (2024). An LLM-based Approach to Recover Traceability Links between Security Requirements and Goal Models. In Proceedings of the 28th International Conference on Evaluation and Assessment in Software Engineering (pp. 643-651). Association for Computing Machinery.

\end{thebibliography}

\end{document}